\documentclass{IEEEtran}

\usepackage{scalerel}
\usepackage{tikz}
\usepackage{amsmath}
\usetikzlibrary{svg.path}
\usepackage{colortbl}
\usepackage{graphicx}
\usepackage{subcaption}
\usepackage{multirow}
\usepackage{units}
\usepackage{framed}
\usepackage{paralist}
\usepackage{hyperref}

\definecolor{orcidlogocol}{HTML}{A6CE39}
\tikzset{
  orcidlogo/.pic={
    \fill[orcidlogocol] svg{M256,128c0,70.7-57.3,128-128,128C57.3,256,0,198.7,0,128C0,57.3,57.3,0,128,0C198.7,0,256,57.3,256,128z};
    \fill[white] svg{M86.3,186.2H70.9V79.1h15.4v48.4V186.2z}
                 svg{M108.9,79.1h41.6c39.6,0,57,28.3,57,53.6c0,27.5-21.5,53.6-56.8,53.6h-41.8V79.1z M124.3,172.4h24.5c34.9,0,42.9-26.5,42.9-39.7c0-21.5-13.7-39.7-43.7-39.7h-23.7V172.4z}
                 svg{M88.7,56.8c0,5.5-4.5,10.1-10.1,10.1c-5.6,0-10.1-4.6-10.1-10.1c0-5.6,4.5-10.1,10.1-10.1C84.2,46.7,88.7,51.3,88.7,56.8z};
  }
}

\newcommand\orcidicon[1]{\href{https://orcid.org/#1}{\mbox{\scalerel*{
\begin{tikzpicture}[yscale=-1,transform shape]
\pic{orcidlogo};
\end{tikzpicture}
}{|}}}}

\title{Direct and Indirect Communication\\
in Multi-Human Multi-Robot Interaction}

\author{
Jayam~Patel\orcidicon{0000-0002-0687-4169}~\IEEEmembership{Student Member, IEEE},
Tyagaraja~Ramaswamy,
Zhi~Li, and
Carlo~Pinciroli\orcidicon{0000-0002-2155-0445}~\IEEEmembership{Member, IEEE}%
\thanks{The authors are with the Department of Robotics Engineering, Worcester Polytechnic Institute, MA, USA. Email: \texttt{\{jupatel, tramaswamy, zli11, cpinciroli\}@wpi.edu}}}

\begin{document}

\maketitle


\begin{abstract}
How can \textit{multiple} humans interact with multiple robots? The goal of our research is to create an effective interface that allows multiple operators to collaboratively control teams of robots in complex tasks. In this paper, we focus on a key aspect that affects our exploration of the design space of human-robot interfaces --- inter-human communication. More specifically, we study the impact of \textit{direct} and \textit{indirect} communication on several metrics, such as awareness, workload, trust, and interface usability. In our experiments, the participants can engage directly through verbal communication, or indirectly by representing their actions and intentions through our interface. We report the results of a user study based on a collective transport task involving 18 human subjects and 9 robots. Our study suggests that combining both direct and indirect communication is the best approach for effective multi-human / multi-robot interaction.
\end{abstract}


\IEEEpeerreviewmaketitle

\section{Introduction}
Multi-robot systems are envisioned in scenarios in which parallelism, scalability, and resilience are key features for success. Scenarios such as firefighting, search-and-rescue, construction, and planetary exploration are typically imagined with teams of robots acting intelligently and autonomously \cite{Brambilla2013}. However, autonomy is only part of the picture — human supervision remains necessary to assess the correct progress of the mission, modify goals, and intervene in case of unexpected faults that cannot be handled autonomously \cite{kolling_human_2016}.

Significant work has been devoted to interfaces that allow \emph{single} human operators to interact with multiple robots. These interfaces typically enable intuitive interaction for specific tasks, such as navigation \cite{ayanianControllingTeamRobots2014}, dispersion \cite{kapellmann-zafraHumanRobotSwarmInteraction2016}, and foraging \cite{kolling_human_2013}. As the complexity of the tasks involved increases, however, the amount of information, the number of robots, and the nature of the interactions is likely to exceed the span of apprehension of any individual operator \cite{lewis2010choosing}. For this reason, it is reasonable to envision that, in future missions with multi-robot systems, \emph{multiple} operators will be involved.

One immediate advantage of multiple operators is the opportunity to partition a large autonomous system in smaller, more manageable parts, each assigned to a dedicated operator. A more interesting insight is that cooperative supervisory control can exploit individual differences among operators. These include diversity in cognitive abilities and in the way operators interact with automation. In this context, the main factors are
\begin{inparaenum}[\it (i)]
\item the ability to focus and shift attention flexibly,
\item the ability to process spatial information, and
\item prior experience with gaming interfaces \cite{chenHumanAgentTeaming2014}. Cooperative supervision that embraces these factors will be extremely effective.
\end{inparaenum}

However, the presence of multiple operators introduces new and interesting challenges. These include ineffective cooperation among the operators \cite{allen2004exploring}, unbalanced workload \cite{mcbride2011understanding,chen2014human}, and inhomogeneous awareness \cite{riley2005situation,lee2008review,parasuraman1997humans}. Neglecting these challenges produces an undesirable phenomenon: the \emph{out-of-the-loop (OOTL) performance problem}, caused by a lack of engagement in the task at hand, of awareness of its state, and of trust in the system and other operators \cite{endsley1995out,gouraud2017autopilot}.

In this paper, we study the impact of inter-human communication on several metrics, such as awareness, workload, trust, and interface usability. Although inter-human communication has been studied extensively~\cite{tomasello2010origins}, it has sparsely been investigated in the context of multi-robot systems, and, to the best of our knowledge, no study exists that focuses on \emph{mobile} teams of robots.

For the purposes of our work, we categorize communication in two broad types: \emph{direct} and \emph{indirect}. In our experiments, direct communication includes verbal and gesture-based modalities, and we define it as an explicit exchange of information among human operators. In contrast, we define indirect communication as an exchange of information that occurs implicitly, through the mediation of, e.g., a graphical user interface. Which elements of the graphical user interface foster efficient indirect communication is a key research question we study in this paper.

This work offers two main contributions:
\begin{enumerate}
\item From the technological point of view, we present a graphical user interface for multi-robot control, with features designed to support information-rich indirect communication among human operators. To the best of our knowledge, this interface is the first of its kind, in that it enables effective interaction between multiple humans and multiple mobile robots in a complex scenario;
\item From the scientific point of view, we study the impact that direct and indirect communication have both in the design of novel interfaces and in the performance of the interaction these interfaces produce.
\end{enumerate}
Our experimental evaluation is based on a study that involved 18 operators, grouped in teams of 2, and 9 real mobile robots involved in an object transport scenario. This paper takes advantage of our previous work on multi-granularity \cite{patel2019} interface design, in which we showed which interaction modalities \cite{Patel:ROMAN2020} and which graphical elements of a user interface \cite{patel2021transparency} provide the best performance in a multi-operator setting.

The paper is organized as follows. In Sec.~\ref{sec-communication:background} we discuss relevant literature on human-robot communication. In Sec.~\ref{sec-communication:framework} we present the design of our interface. In Sec.~\ref{sec-communication:userstudy} we introduce the user study, and discuss our main findings in Sec.~\ref{sec-communication:discussion}. We conclude the paper in Sec.~\ref{sec-communication:conclusion}.

\section{Background}
\label{sec-communication:background}

\textbf{Granularity of control.}
Over the last two decades, HRI research on mobile multi-robot systems has focused on identifying suitable interfaces and investigating methods to effectively interact the robots. One of the key aspects of these interfaces is control granularity~\cite{kolling_human_2016,endsley2017here}, and it generally includes robot-oriented, team-oriented, and environment-oriented control. In robot-oriented control, the operator interacts with a single robot, either statically predetermined or dynamically selected~\cite{setter_haptic_2015, kapellmann-zafra_human-robot_2016, nagi_human-swarm_2014, alonso-mora_gesture_2015, gromov_wearable_2016}. Team-oriented control allows an operator to assign a goal to a groups of robots as if the group were a single entity~\cite{natraj_gesturing_2014, ayanian_controlling_2014, diaz-mercado_distributed_2015}. Environment-oriented control occurs when the operators modifies the environment, typically using augmented or mixed reality, to influence the behavior of the robots or to indicate goals~\cite{bashyal_human_2008, kolling_human_2016, patel2019,Patel:ROMAN2020,patel2021transparency}. These control modalities are typically studied with a \emph{single} operator in mind.

\textbf{Human-robot communication.}
Research on communication has mostly focused on the relationship between a human operator and one or more robots. The distinction we consider in this paper between direct and indirect communication is a common aspect of human-robot communication modalities~\cite{mavridis2015review, saunderson2019robots}. Notable works include direct human-robot communication through natural language conveyed verbally~\cite{bisk2016natural} and non-verbal communication through social cues~\cite{sartorato2017improving}, facial expressions~\cite{liu2017facial} and eye gaze~\cite{admoni2017social, breazeal2005effects}. Lakhmani \emph{et al.} \cite{lakhmani_exploring_2019} presented a method for direct communication between an operator and a robot in which communication can be either \textit{directional} or \textit{bidirectional}. In directional communication, only the robot can send information to the operator. In bidirectional communication, the operator and the robot can send information to each other. The authors reported that the operator performed better with directional communication as compared to bidirectional communication. Indirect communication has been studied in several contexts, including:
\begin{inparaenum}[\it (i)]
\emph{(i)} a human operator attempting to infer the behavior of the robots~\cite{knight_expressive_2016, capelli_communication_2019}, and
\emph{(ii)} robots attempting to predict the actions of the \emph{human-in-the-loop}~\cite{claes_multi_2018,wang_trust-based_2018,bajcsy_scalable_2018,zhang_optimal_2016}.
\end{inparaenum}
Che \emph{et al.}~\cite{che2020efficient}, compare the effects of direct and indirect communication between a human and a robot in a navigation task. In this work, the human acts as a bystander and the robot must navigate around the human. The robot can either indirectly predict the human’s direction of movement and navigate around, or can directly notify the human about its intentions of moving in a direction. The authors reported that the combination of indirect and direct communication positively impacts the humans performance and trust.

\textbf{Conflicts among multiple operators.}
To the best of our knowledge, analogous studies in communication that focus on multiple human operators are currently missing. The most notable works in multi-human multi-robot interaction concern inter-operator conflict in scenarios in which the operators do not communicate. A conflict might arise when multiple operators wish to control the same robot or specify incompatible goals. If communication is not possible, pre-assigning robots among operators is a possible solution. Lee \emph{et al.}~\cite{lee_teams_2010} compare the performance of two scenarios: in one, the operators control robots from a shared pool of robots; and the other, the operators have disjoint assigned pools. The operators can engage with robots in a robot-oriented fashion and manipulate one robot at a time. Lee \emph{et al.}’s findings show that the performance of the operators is better when they can manipulate robots from the assigned pools. Their performance drops when they control robots from the shared pool, due to the risk of specifying conflicting goals for the same robots. In a similar vein, You \emph{et al.}~\cite{you_curiosity_2016} study a scenario in which two operators control separate robots, and with them push physical objects from one place to another. You \emph{et al.} divide the environment into two regions, with a operator-robot pair assigned to each region and limited the operator from moving from one region to another. No conflicts are possible in this case.

\textbf{Novelty.}
In this paper, we take inspiration from the three research strands discussed above, and combine them in a coherent study. We investigate whether an interface that offers multiple granularity of control can enable the operators to decide when and how to control the robots, and study the role that communication has in the design of such interface and in the way the operators collaborate. To the best of our knowledge, this study is the first to study these research questions together.

\section{Multi-Human Multi-Robot Interaction System}
\label{sec-communication:framework}
\subsection{System Overview}


\begin{figure}[t]
    \centering
    \includegraphics[width=\linewidth]{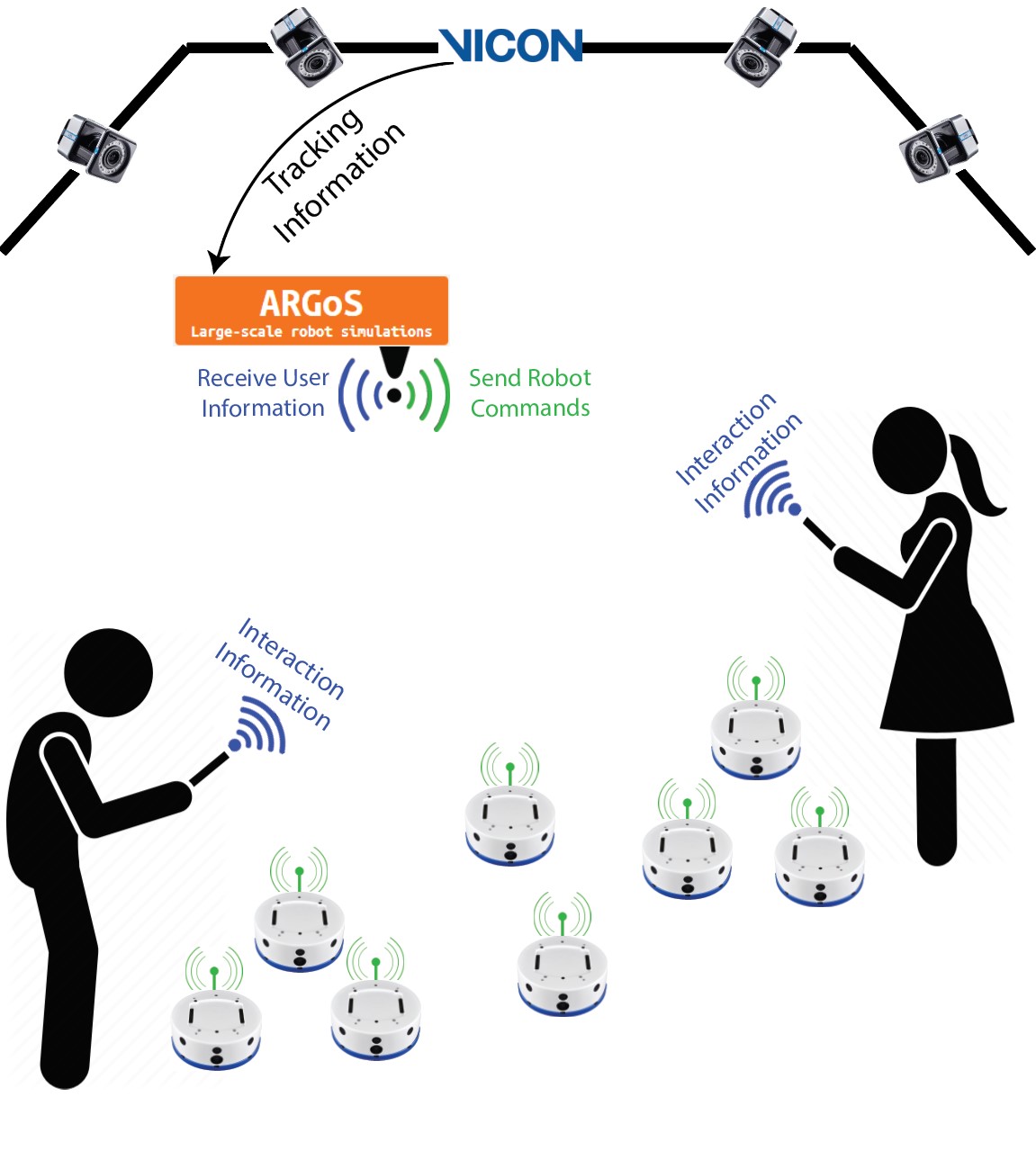}
    \caption{System overview.}
    \label{fig-communication:system_overview}
\end{figure}

Our overall system, schematized in Fig.~\ref{fig-communication:system_overview}, consists of a mixed-reality (MR) interface, a team of 9 Khepera IV robots,\footnote{\url{https://www.k-team.com/khepera-iv}} a Vicon motion tracking system,\footnote{\url{http://vicon.com}} and ARGoS~\cite{Pinciroli:SI2012}, a fast and modular multi-robot simulator. The MR interface enables an operator to interact with the robots, displaying useful information on actions and intentions of robots and other operators. The Khepera IV are differential-drive robots that we programmed to perform behaviors such as navigation and collective object transport. ARGoS acts as the software glue between the MR interfaces, the robots, and the Vicon.

To this aim, we modified ARGoS by replacing its simulated physics engine with a new plug-in that receives positional data from the motion capture systems. In addition, we created plug-ins that allow ARGoS to communicate directly with the robots for control and debugging purposes. These plug-ins create a feedback loop between the robots and the environment which allows us to programmatically control what the robots do and sense as an experiment is running. Thanks to this feature of our setup, the MR interfaces, once connected to ARGoS, can both retrieve data and convey modifications made using the control modalities offered by the interfaces. To make this possible, ARGoS also translates the coordinates seen by the MR interface with those seen by the Vicon and viceversa.

When an operator defines a new goal position for the robots or the objects present in the system, the MR interface transmits the goal to ARGoS. The latter converts the request into motion primitives and transmits them to the robots. Finally, the robots execute the primitives. At any time during the execution, the operators can interfere with the system by defining new goals for the robots. This feature is particularly important when robots temporarily misbehave due to faults or noise.

\subsection{Operator Interface and Control Modalities}

\begin{figure}[t]
    \centering
    \includegraphics[width=0.49\textwidth]{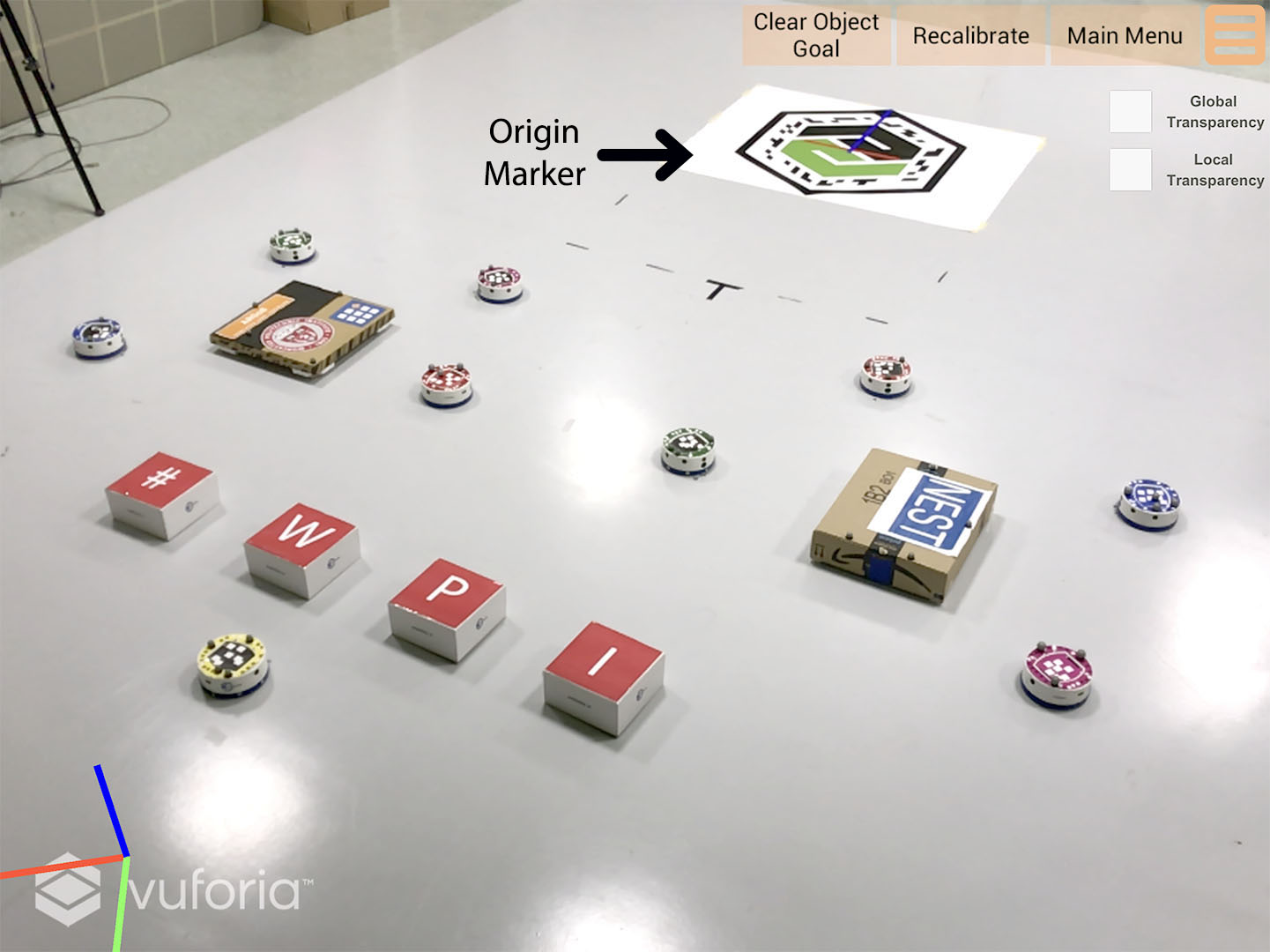}
    \caption{Screenshot of the MR interface running on an iPad. The black arrow was added in this picture to indicate the position of the origin marker. This marker is used by Vuforia to identify the origin of the environment.}
    \label{fig-communication:screenshot}
\end{figure}

The interaction between the operators and the robots happens through the MR interface installed on an Apple iPad. We implemented the MR interface with Vuforia,\footnote{\url{http://vuforia.com}} a software development kit for mixed-reality applications. Vuforia provides a functionality to recognize and track physical objects with fiducial markers. We integrated Vuforia with the Unity Game Engine\footnote{\url{http://unity3d.com}} for visualization. Fig.~\ref{fig-communication:screenshot} shows a screenshot of the view through the app running on an iPad. 

The MR interface recognizes objects and robots through unique fiducial markers applied to the each entity of interest. The interface overlays virtual objects on the recognized fiducial markers. The operator can move these virtual objects using a one-finger swipe and rotate them with a two-finger twist. The operator can also select multiple robots (for, e.g., collective motion) by drawing a closed contour. The manipulation of robots and objects occurs in three phase: start, move, and end. The \emph{start} phase is initiated when the operator touches the handheld device. Once the touch is detected, the \emph{move} phase is executed as long as the operator is performing a continuous motion gesture on the device. The \emph{end} phase is triggered when the operator releases the touch. After the completion of the \emph{end} phase, the interface parses the gesture and acts accordingly, e.g., sends the final pose of a selected object to ARGoS. The interface enables three interaction modalities: object-oriented (a special case of the environment-oriented modality), robot-oriented, and team-oriented, explained in the rest of this section.

\begin{figure}[t]
  \centering
  \begin{subfigure}[t]{0.23\textwidth}
    \includegraphics[width=\textwidth]{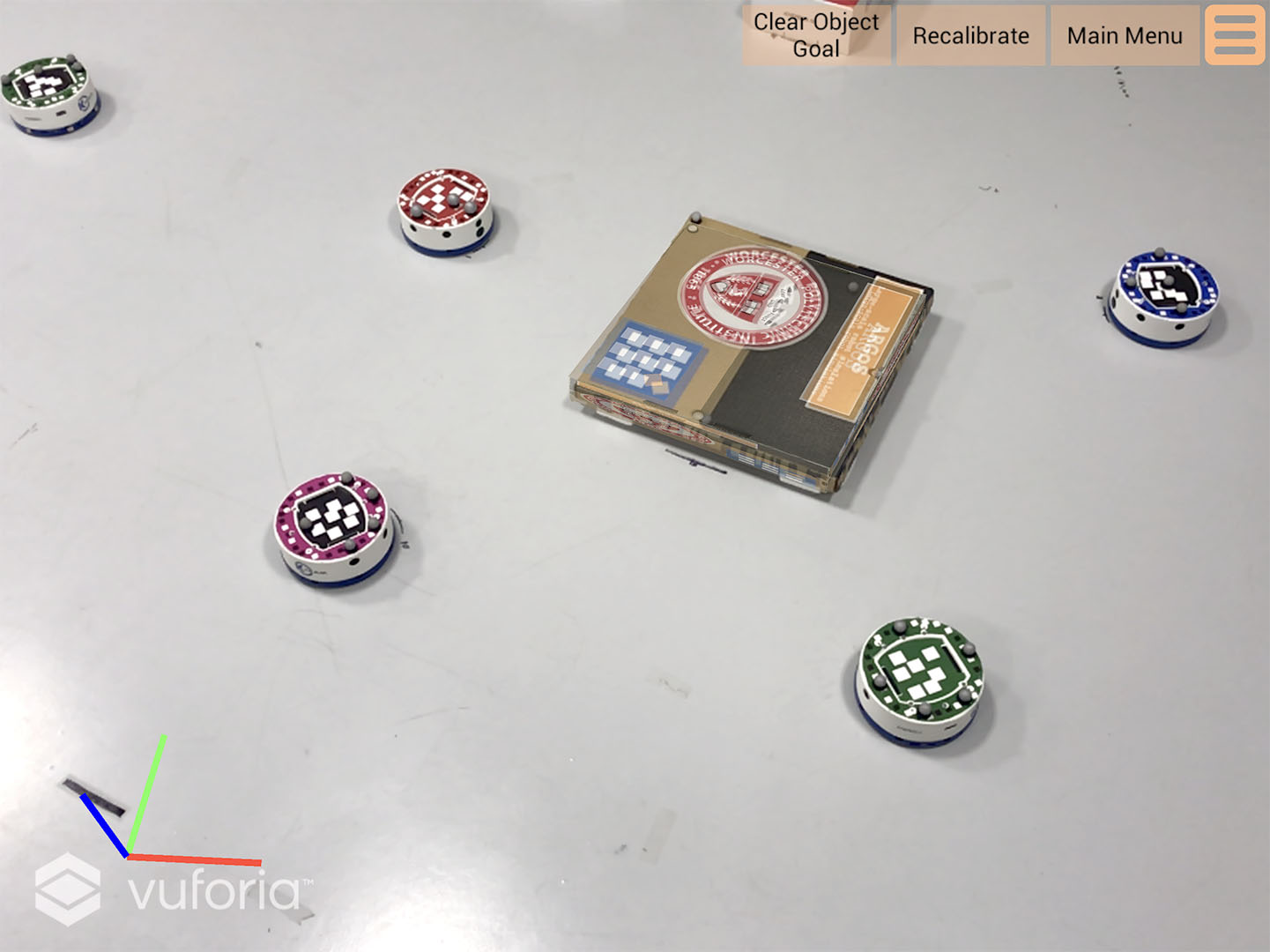}
    \caption{Object recognition}
    \label{fig-communication:modeO1}
  \end{subfigure}
  \begin{subfigure}[t]{0.23\textwidth}
    \includegraphics[width=\textwidth]{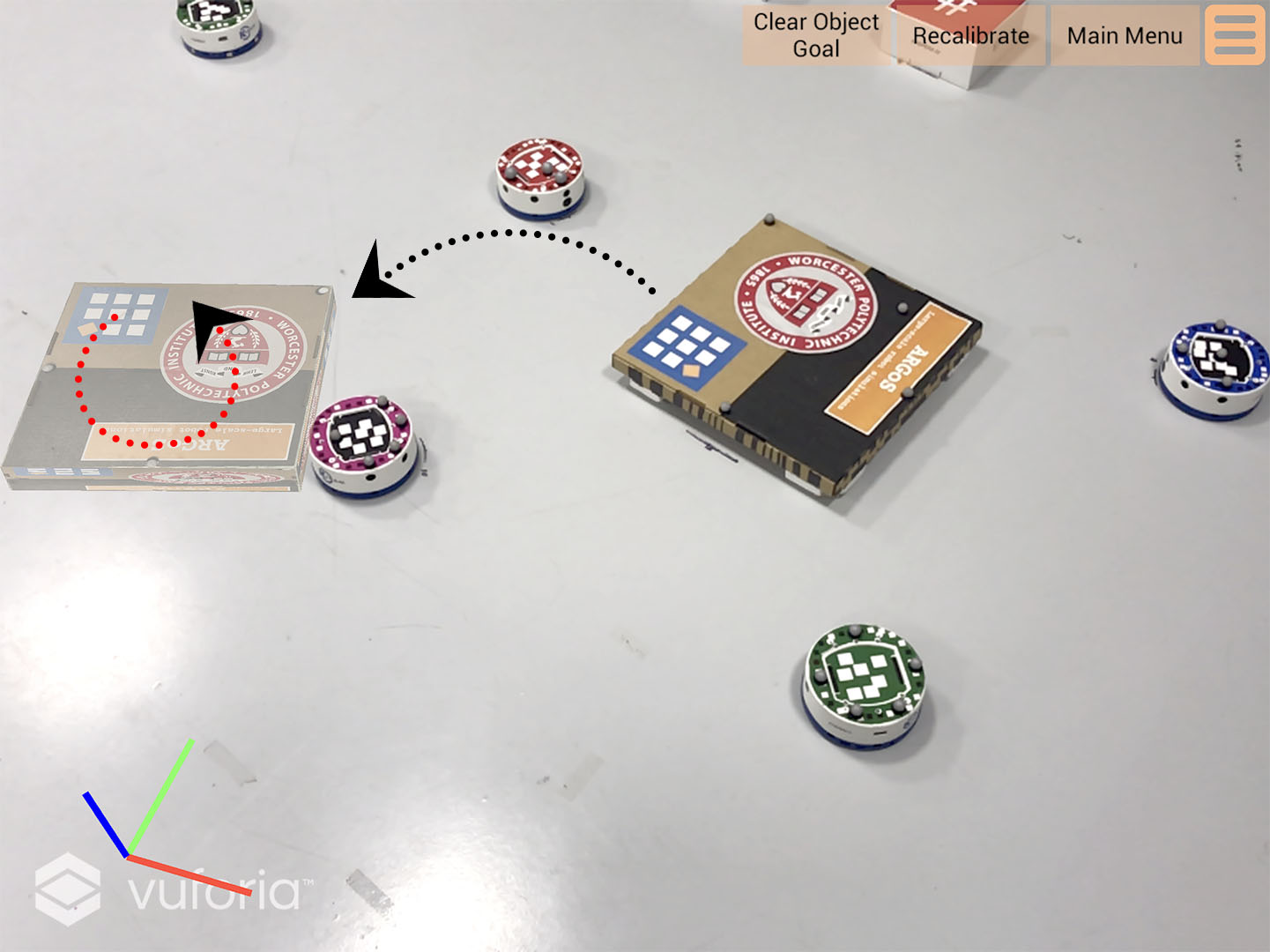}
    \caption{New Goal Defined}
    \label{fig-communication:modeO2}
  \end{subfigure}
  \begin{subfigure}[t]{0.23\textwidth}
    \includegraphics[width=\textwidth]{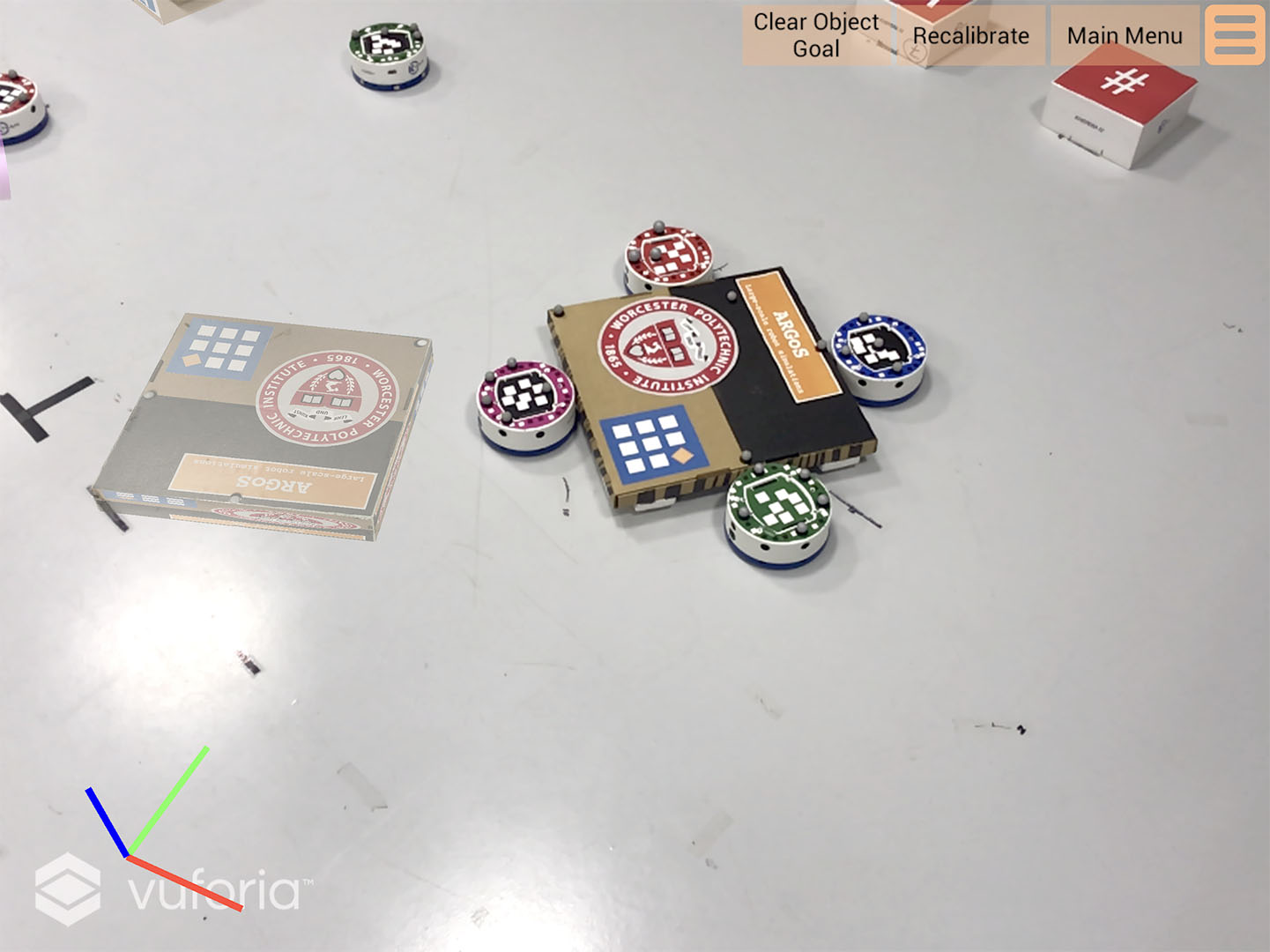}
    \caption{Robots approach and push}
    \label{fig-communication:modeO3}
  \end{subfigure}
  \begin{subfigure}[t]{0.23\textwidth}
    \includegraphics[width=\textwidth]{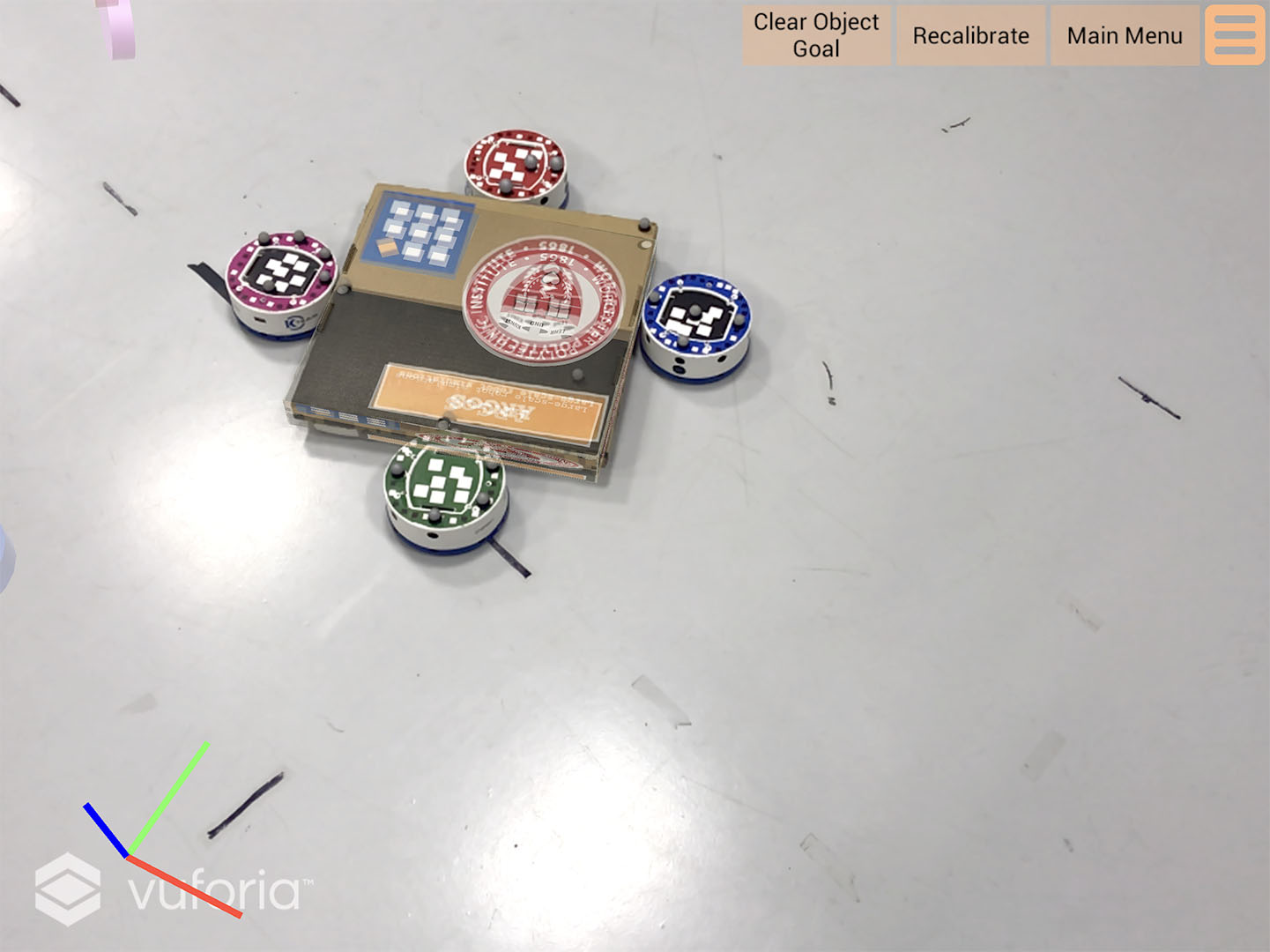}
    \caption{Transport complete}
    \label{fig-communication:modeO4}
  \end{subfigure}
  \caption{Object manipulation by interaction with the virtual object through the interface. The dotted black arrow indicates the one-finger swipe gesture used to move the virtual object and the red dotted arrow indicates the two-finger rotation gesture.}\label{fig-communication:modeO}
\end{figure}

\textbf{Object-oriented Interaction.} The interface overlays a virtual cuboid on objects equipped with a fiducial marker (see Fig.~\ref{fig-communication:modeO1}). The dimensions and textures of the virtual objects match the dimensions and textures of the fiducial markers. The operators can differentiate between virtual cuboids using these similarities while simultaneously moving multiple objects. The operators can move these virtual objects using a one-finger swipe and rotate it with a two-finger twist (see Fig.~\ref{fig-communication:modeO2}). If two or more operators simultaneously want to move the same virtual object, then the robots transport the object to the last received position. Fig.~\ref{fig-communication:modeO3} and Fig.~\ref{fig-communication:modeO4} show the robots transporting the object.

\begin{figure}[t]
  \centering
  \begin{subfigure}[t]{0.23\textwidth}
    \includegraphics[width=\textwidth]{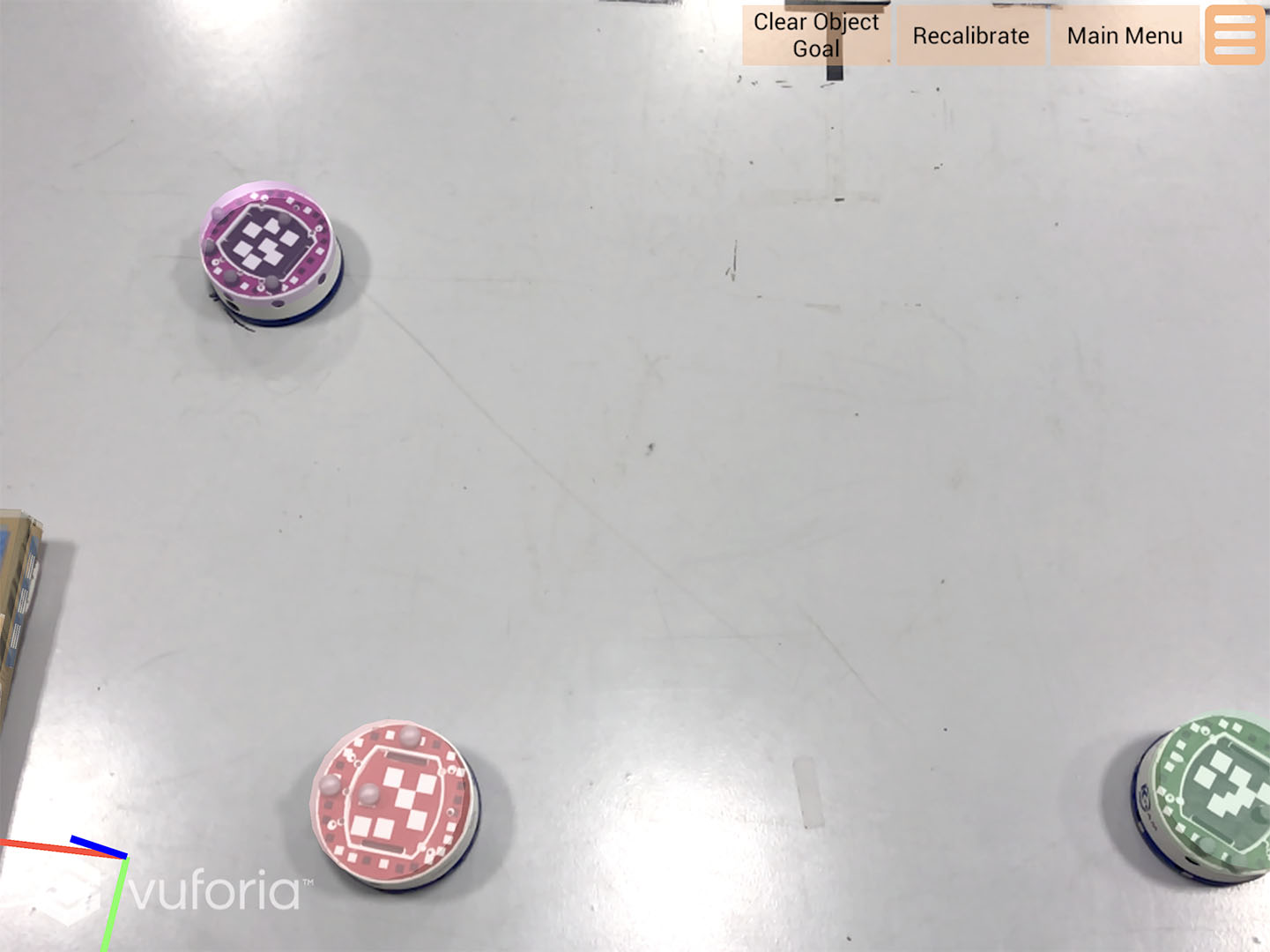}
    \caption{Robot recognition}
    \label{fig-communication:modeR1}
  \end{subfigure}
  \begin{subfigure}[t]{0.23\textwidth}
    \includegraphics[width=\textwidth]{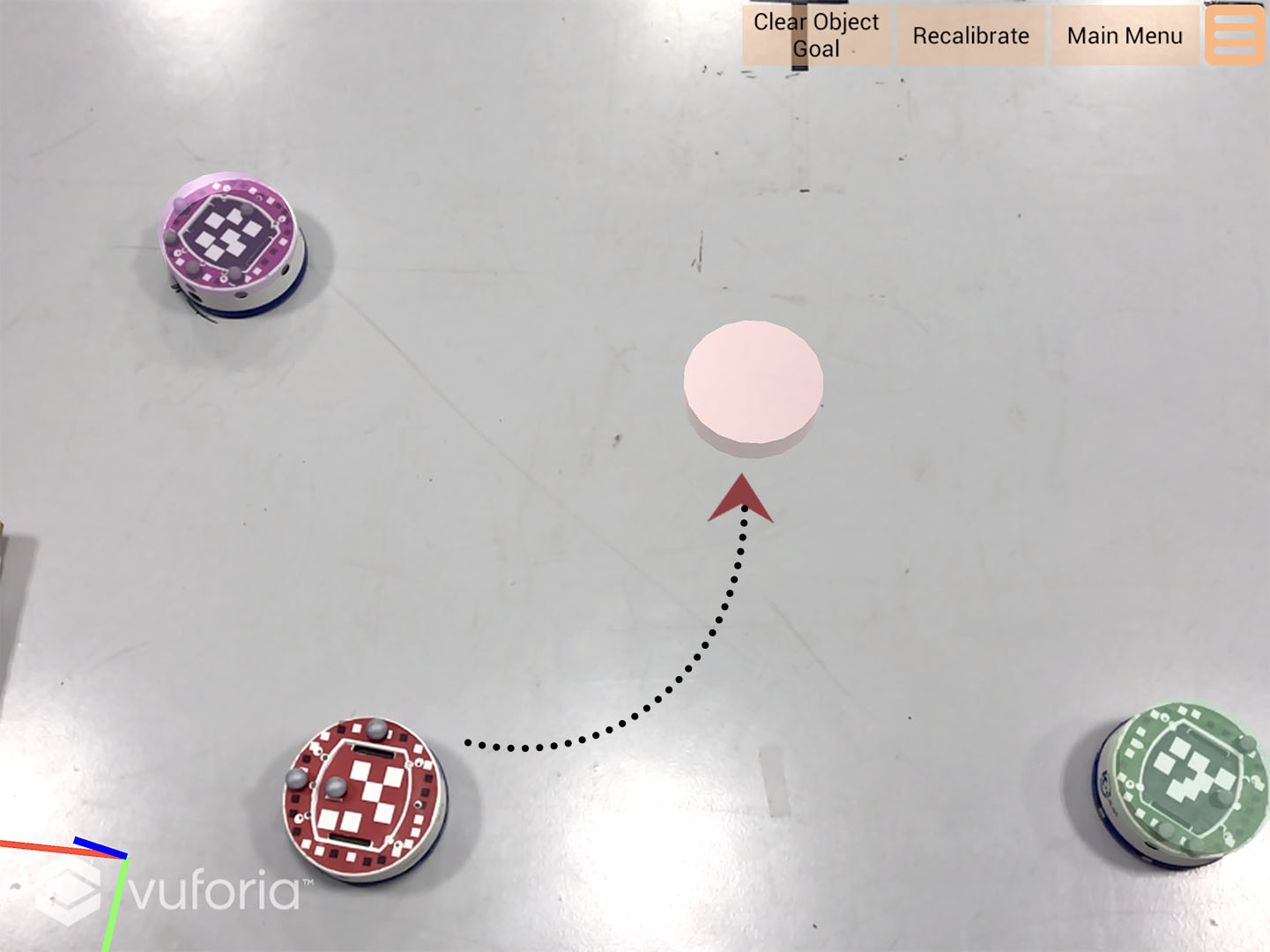}
    \caption{New robot position}
    \label{fig-communication:modeR2}
  \end{subfigure}
  \caption{Robot manipulation by interaction with the virtual robots through the interface. The dotted black arrow indicates the one-finger swipe gesture to move the virtual robot and the arrowhead color indicates the moved virtual robot.}\label{fig-communication:modeR}
\end{figure} 

\textbf{Robot-oriented Interaction.} The interface overlays a virtual cylinder on the recognized robot tags (see Fig.~\ref{fig-communication:modeR1}). The dimensions and colors of the virtual cylinder are identical to the dimensions and colors of the fiducial marker on the robot. Operators can differentiate between different virtual cylinders using these colors while controlling multiple robots at once. The operators can use a one-finger swipe to move the virtual cylinders to denote a desired position for the robots (see Fig.~\ref{fig-communication:modeR2}). If multiple operators simultaneously want to move the virtual cylinder, then the robot considers the last received request. If the robot is performing collective object transport, then other robots in the team pause their operation until the selected robot reaches the desired position. If the robot is part of an operator-defined team, then the other robots are not affected by the position change. The robot control can also be used to manually push the object.

\begin{figure}[t]
  \centering
  \begin{subfigure}[t]{0.23\textwidth}
    \includegraphics[width=\textwidth]{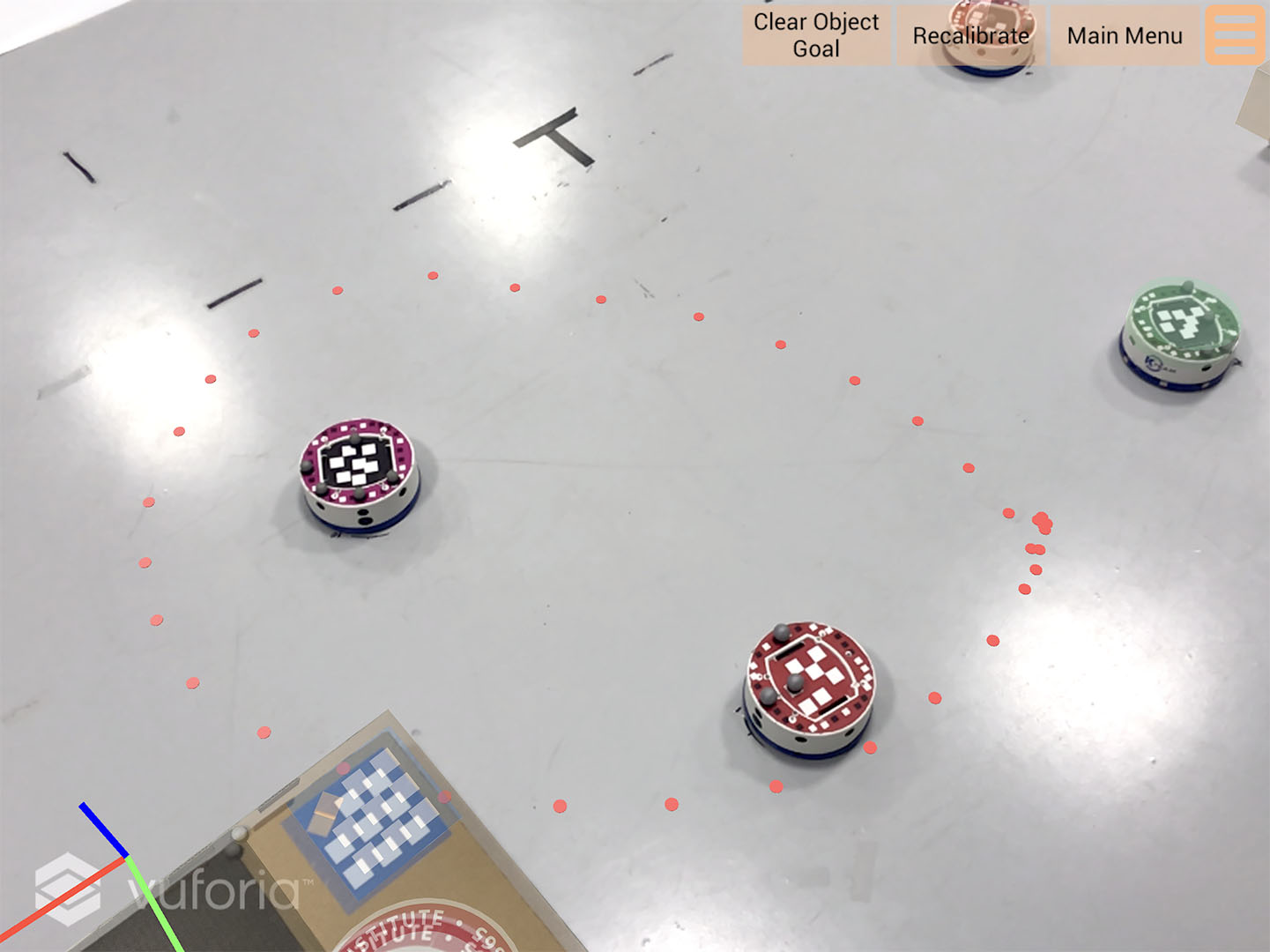}
    \caption{Robot team selection}
    \label{fig-communication:modeS1}
  \end{subfigure}
  \begin{subfigure}[t]{0.23\textwidth}
    \includegraphics[width=\textwidth]{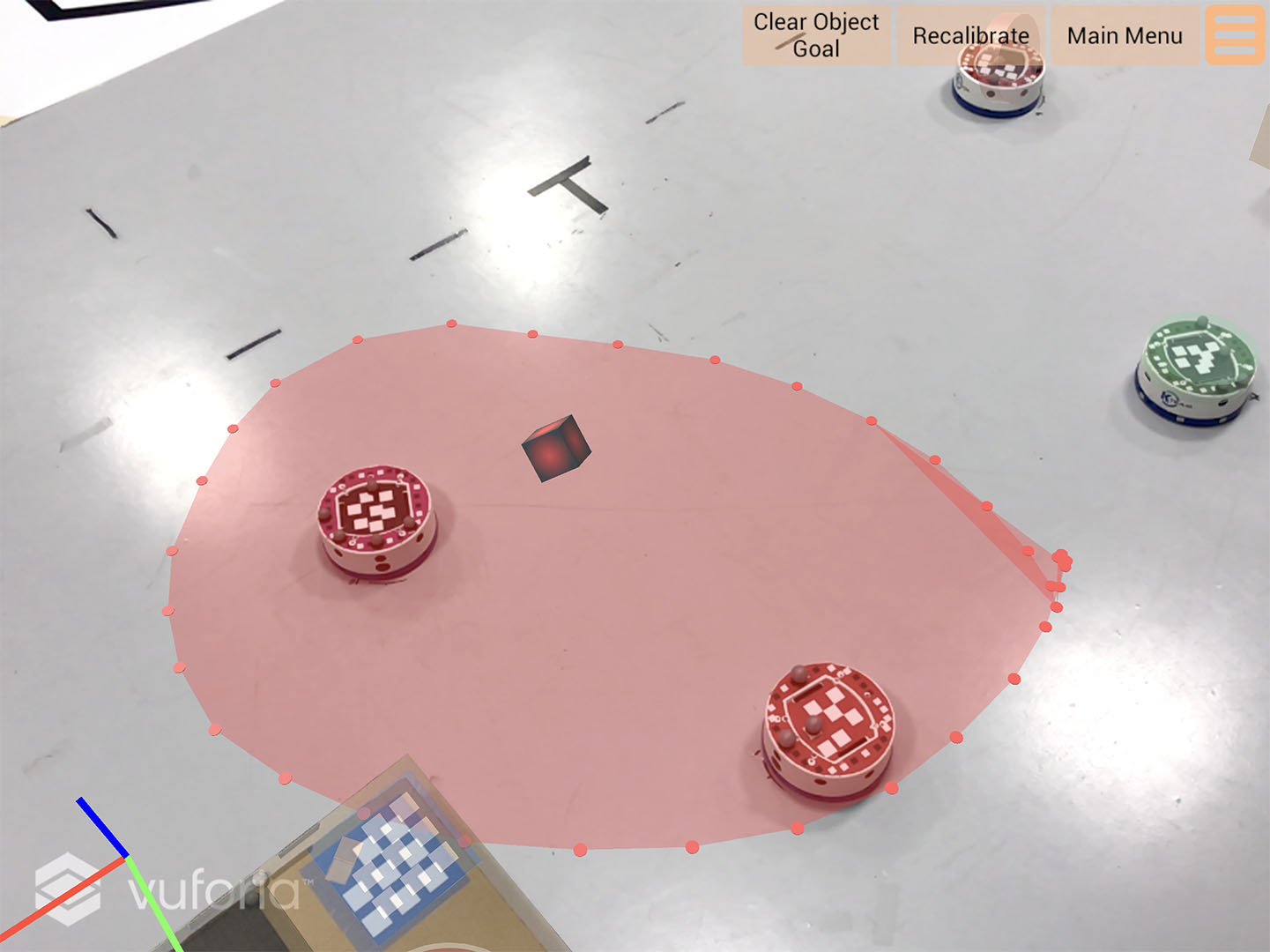}
    \caption{Robot team creation}
    \label{fig-communication:modeS2}
  \end{subfigure}
  \begin{subfigure}[t]{0.23\textwidth}
    \includegraphics[width=\textwidth]{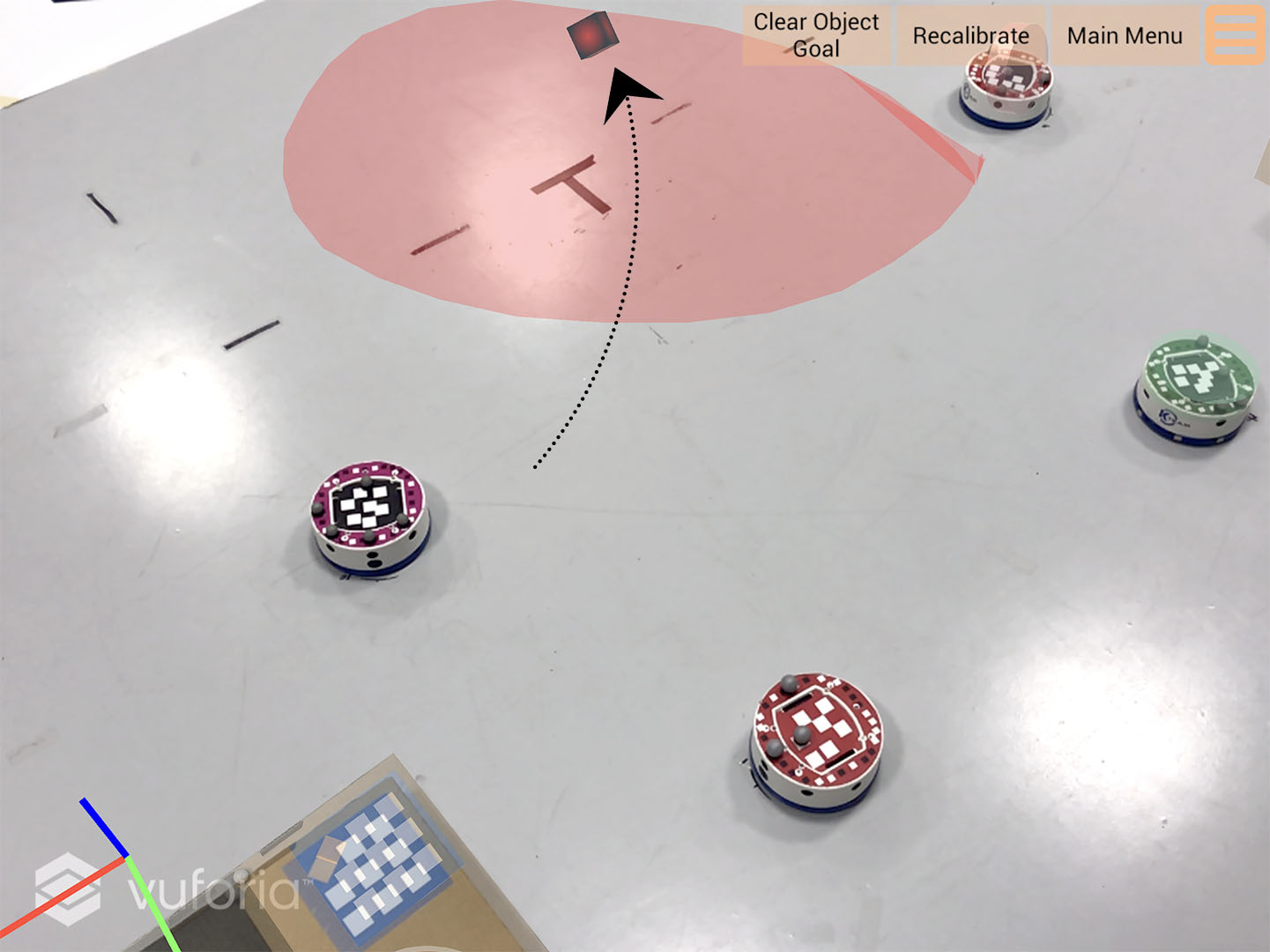}
    \caption{Robot team manipulation}
    \label{fig-communication:modeS3}
  \end{subfigure}
  \begin{subfigure}[t]{0.23\textwidth}
    \includegraphics[width=\textwidth]{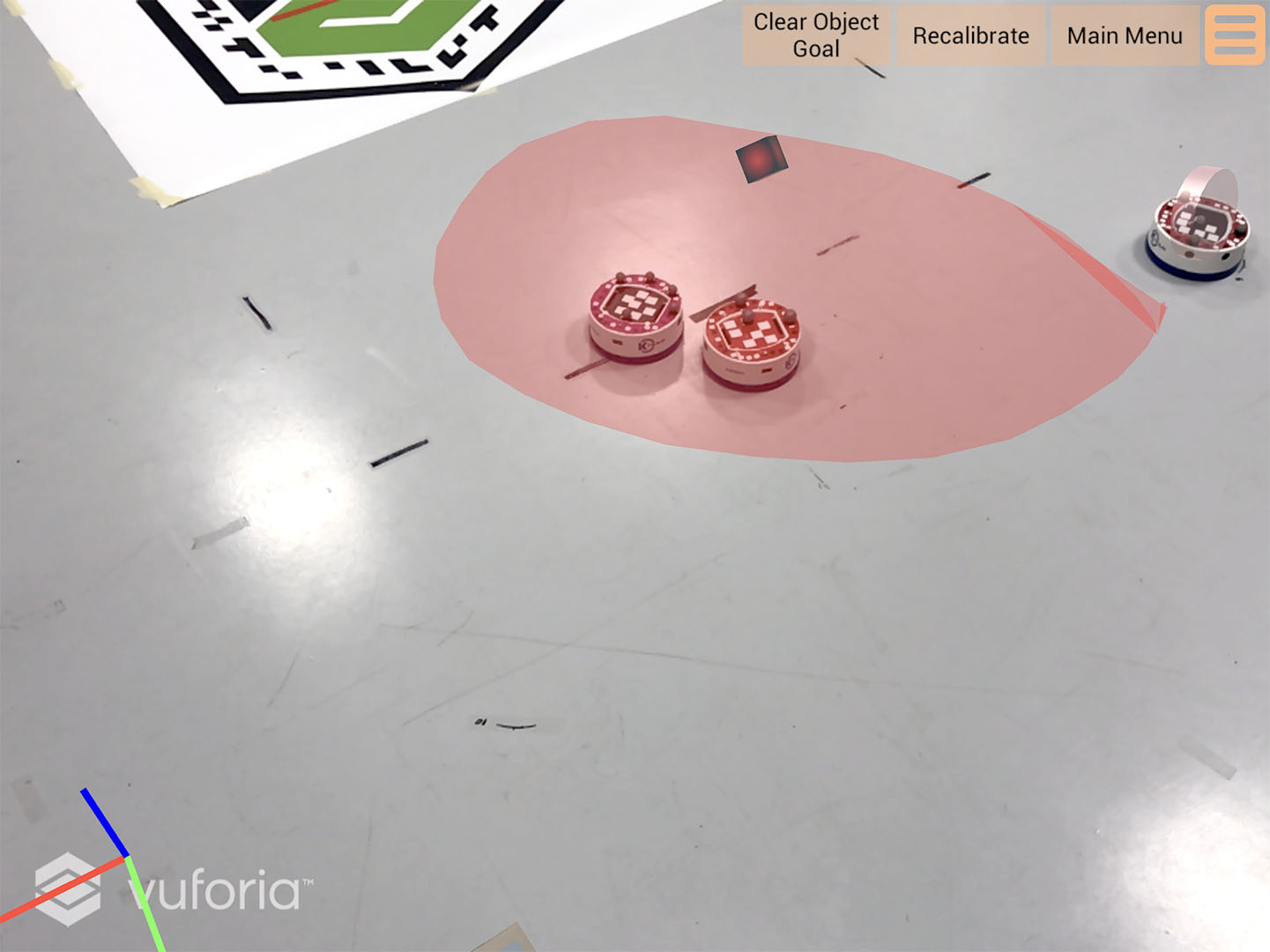}
    \caption{Robot team re-positioned}
    \label{fig-communication:modeS4}
  \end{subfigure}
  \caption{Robot team creation and manipulation. The dotted black arrow indicates the one-finger swipe gesture to move the virtual cube for re-positioning the team of robots.}
  \label{fig-communication:modeS}
\end{figure} 

\textbf{Robot Team Selection and Manipulation.} Operators can draw a closed contour with a one-finger continuous swipe to select robots enclosed in the shape (see Fig.~\ref{fig-communication:modeS1}). The interface draws the closed contour in red color to show the area of selection (see Fig.~\ref{fig-communication:modeS2}). The interface displays a cube, hovering at the centroid of the created contour, to represent the team of robots (see Fig.~\ref{fig-communication:modeS2}). The operators can move this cube with a one-finger swipe to define a desired location for the team to navigate (see Fig.~\ref{fig-communication:modeS4} and Fig.~\ref{fig-communication:modeS4}). Each operator can create only one custom team at a time and the interface deletes all previous selections. If a robot is part of multiple teams defined by multiple operators, then the robot moves to the latest location. 

\subsection{Collective Transport}

\begin{figure}[t]
    \centering
    \includegraphics[width=0.49\textwidth]{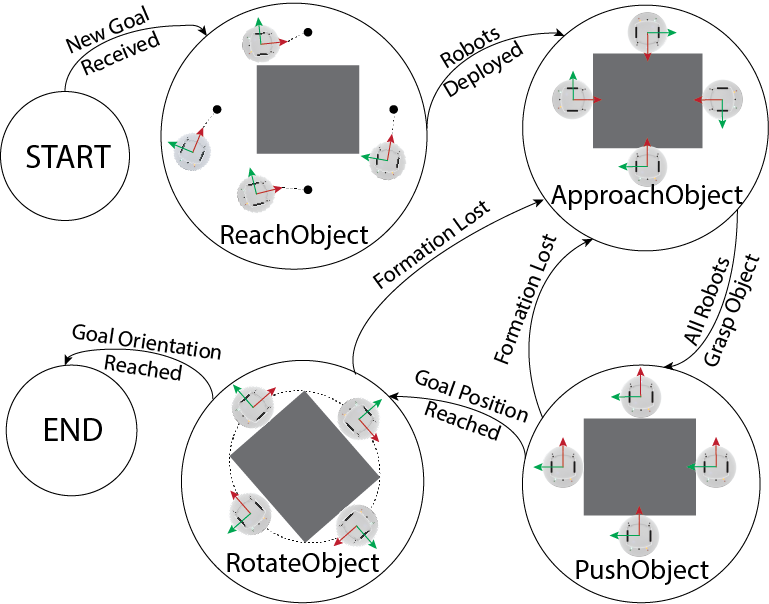}
    \caption{Collective transport state machine.}
    \label{fig-communication:collective_transport}
\end{figure}

We used a finite state machine to implement our collective transport behaviour. The finite state machine is executed independently by each robot, using information from the Vicon and the MR mediated by ARGoS. Fig.~\ref{fig-communication:collective_transport} shows the state machine, along with the conditions for state transition. We define the states as follows:

\textbf{Reach Object.} The robots navigate to the object upon receiving a desired position from the operator. The robots arrange themselves around the object. The state ends when all the robots reach their chosen positions.   

\textbf{Approach Object.} The robots face the centroid of the object and start moving towards it. The state ends when all the robots are in contact with the object. 

\textbf{Push Object.} The robots orient themselves in the direction of the goal position and start moving when all robots are facing that direction. The robot in the front of the formation maintains a fixed distance from the object allowing all the robots to stay in the formation. The state transitions to \emph{Approach Object} if the formation breaks. The state ends when the object reaches the desired position. 

\textbf{Rotate Object.} The robots arrange themselves around the object and move in a circular direction to rotate the object. The state transitions to \emph{Approach Object} if the formation breaks. The state ends when the object achieves the desired orientation.

\subsection{Information Transparency}

\begin{figure}[t]
    \centering
    \includegraphics[width=\linewidth]{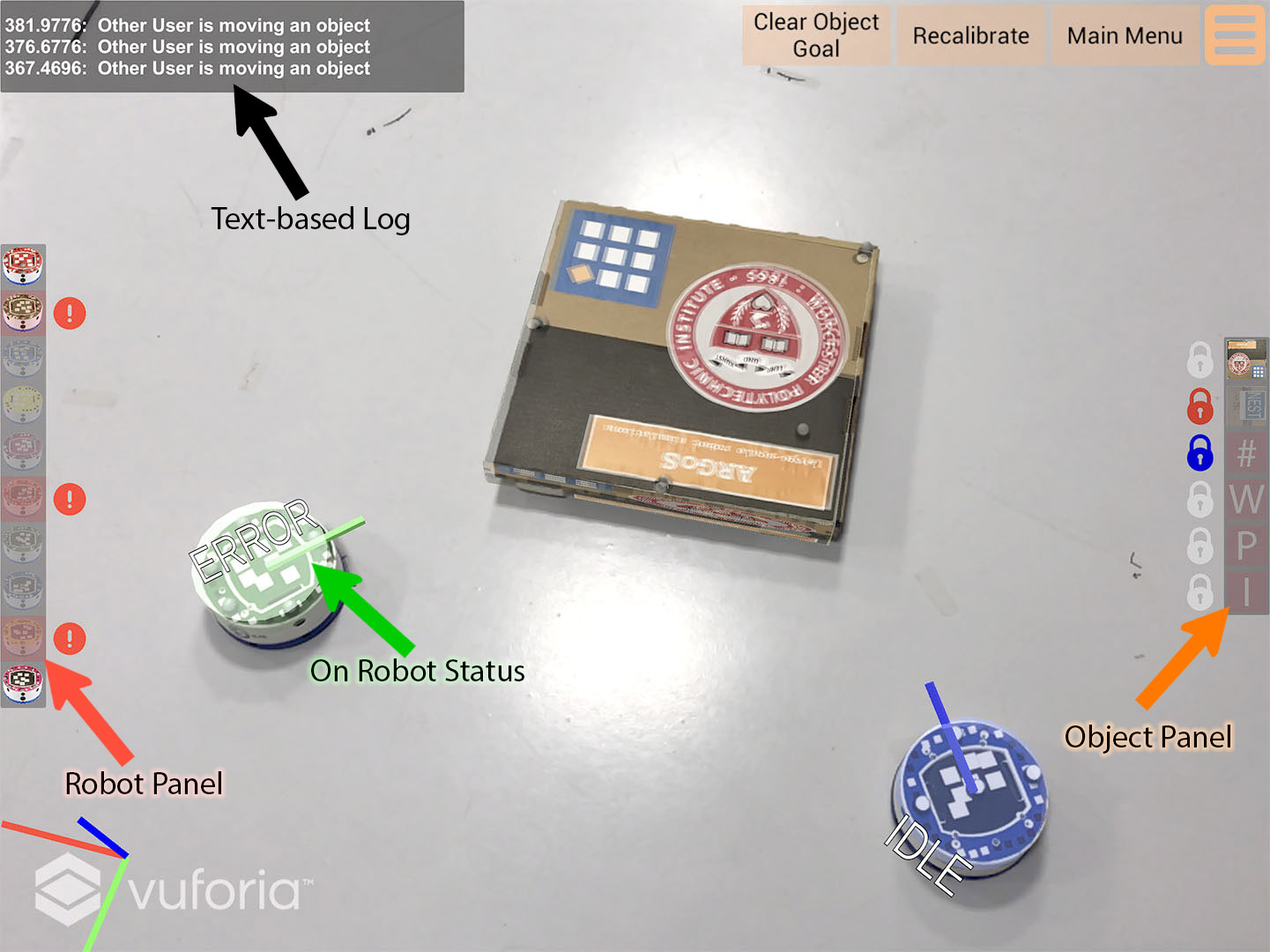}
    \caption{Transparency features in the mixed-reality interface. The black arrow indicates the text-based log. The red arrow indicates the Robot Panel. The green arrow indicates the on-robot status. The orange arrow indicates the object panel.}
    \label{fig-communication:transparency}
\end{figure}

\begin{figure}[t]
    \centering
    \includegraphics[width=\linewidth]{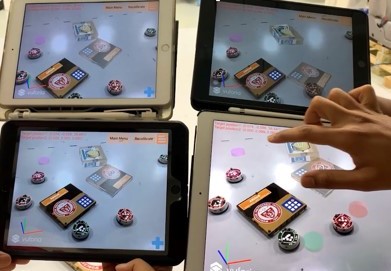}
    \caption{Shared awareness demonstrated using four mixed-reality interfaces.}
    \label{fig-communication:sharedawareness}
\end{figure}


We designed our interface to provide a rich set of data about the robots, the task progress, and other operators’ actions. We studied the effectiveness of these features in previous work~\cite{patel2021transparency}, assessing how they contribute to making the system more \emph{transparent}~\cite{bhaskara_agent_2020} for the operators. The interface is shown in Fig.~\ref{fig-communication:transparency}.

To foster shared awareness, when an operator manipulates a virtual object, the action is broadcast to the other interfaces in real-time, making it possible for every operator to see what is being done by other operators. Fig.~\ref{fig-communication:sharedawareness} shows the image of four mixed-reality interfaces demonstrating this shared awareness feature.

The information is organized into a set of panels, updated in real time whenever robots and operators perform actions that modify the state of the system and progress of the tasks in execution. The interface offers the following information widgets:
\begin{itemize}
\item Small ‘on-robot’ panels that follow each robot in view to convey their current state and actions. These panels are updated when operators assign new tasks to the robots. In case of faults, this panel displays an error message that helps to identify which robot has malfunctioned.
\item A robot panel (on the left side of the screen) that graphically indicates functional and faulty robots. The robots currently engaged in a task are displayed with specific color shades. The panel also displays a blinking red exclamation point in case a robot acts in a faulty manner.
\item A text-based log (on the top-left of the screen) that notifies the operators about other operators’  interactions with robots and objects. The interface updates the log every time other operators manipulate any virtual object. The log stores the last three actions and discards the rest. 
\item An object panel (on the right side of the screen) that provides information regarding the objects that are being manipulated the interface user and the other operators. The panel also highlights the object icon corresponding to the object being moved by the robots. Additionally, the interface offers the option of selecting an object icon to lock it for future use. An operator can lock the object by tapping the lock icon. This changes the color of the lock to blue, signifying the lock. An operator can select only one object at a time. The interface of the other operators highlights the lock with a red icon.
\end{itemize}

\section{User Study}
\label{sec-communication:userstudy}

\subsection{Communication Modes and Hypotheses}
This study aims to investigate the effects of direct and indirect communication on multi-human multi-robot interaction. The consider three modalities of communication: direct, indirect, and mixed (the combination of direct and indirect). We based our experiments on three main hypotheses.
\begin{enumerate}
\item[\textbf{H1:}] Mixed communication (MC) has the best outcome compared to the other communication modalities in terms of situational awareness, trust, interaction score and task load.
\item[\textbf{H2:}] Operators prefer mixed communication (MC) to the other modes.
\item[\textbf{H3:}] Operators prefer direct communication (DC) over indirect communication (IC).
\end{enumerate}

\subsection{Task Description}
\begin{figure}[t]
  \centering
  \includegraphics[width=0.49\textwidth]{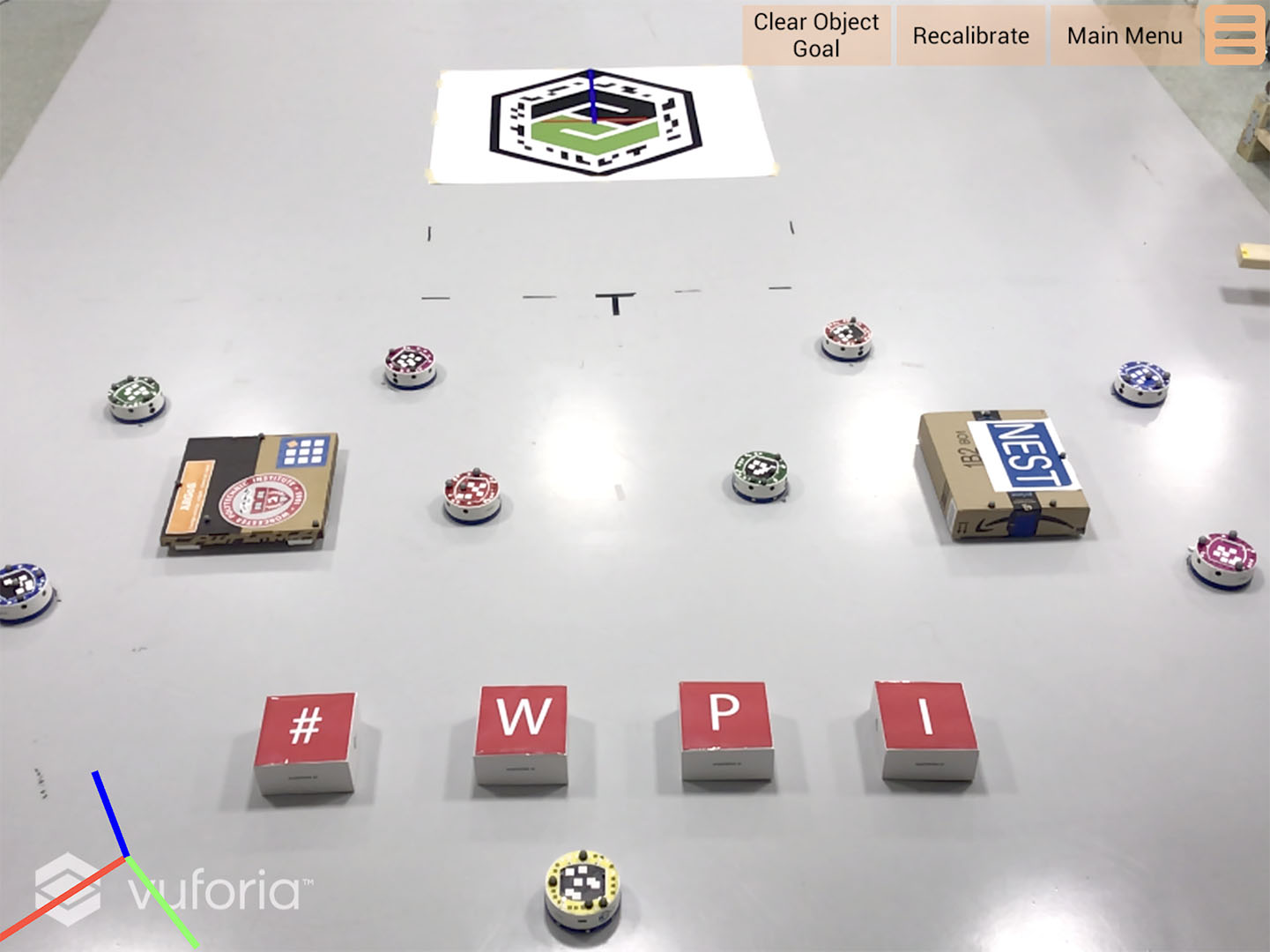}
  \caption{User study experiment setup.}
  \label{fig-communication:study_setup}
\end{figure}
We designed a gamified scenario in which the operators must instruct the robots to transport objects to target areas. The environment had 6 objects (2 big objects and 4 small objects), and the operators received points according to which objects were successfully transported to their target. The big objects were worth 2 points each and the small objects were 1 point each, for a total of 8 points.

The operators, in teams of 2, had a maximum of 8 minutes to accrue as many points as possible. The operators could manipulate big objects with any kind of interaction modality (object-oriented, team-oriented, robot-oriented); small objects, in contrast, could only be manipulated with the robot- and team-oriented modalities. The operators were given 9 robots to complete the game.

Fig.~\ref{fig-communication:study_setup} shows the initial positions of the objects and the robots. All the participants had to perform the task four times with a different communication modalities as follows:

\begin{itemize}
\item \textbf{No Communication (NC):} The operators are not allowed to communicate in any way;
\item \textbf{Direct Communication (DC):} The operators can communicate verbally and non-verbally, but the interfaces do not broadcast information.
\item \textbf{Indirect Communication (IC):} The operators can communicate indirectly through the transparency features of the interface, but are not allowed to talk, use gestures, or infer from body language what other operators are doing;
\item \textbf{Mixed Communication (MC):} The complete set of communication modalities is allowed for both operators.
\end{itemize}

\subsection{Participant Sample}
We recruited 18 students from our university (11 male, 7 females) with ages ranging from 20 to 30 ($M = 24.17$, $SD = 2.68$). No participant had prior experience with our interface, our robots, or even the laboratory layout.

\subsection{Procedures}
Each session lasted approximately 105 minutes. We explained the study to the participants after signing a consent form, and gave the participants 10 minutes to play with the interface. We randomized the order of the tasks in an attempt to reduce the influence of learning effects. 

\subsection{Metrics}
We recorded subjective measures from the operators and objective measures using ARGoS for each game. We used the following scales as metrics:

\begin{figure}[t]
  \centering
  \begin{framed}
    \setdefaultleftmargin{0pt}{}{}{}{}{}
    \begin{compactitem}
    \item Did you understand your \textit{teammate’s intentions}? Were you able to understand why your teammate was taking a certain action?
    \item Could you understand your \textit{teammate’s actions}? Could you understand what your teammate was doing at any particular time?
    \item Could you follow the \textit{progress of the task}? While performing the tasks, were you able to gauge how much of it was pending?
    \item Did you understand what the \textit{robots were doing}? At all times, were you sure how and why the robots were behaving the way they did?
    \item Was the information provided by the interface \textit{clear to understand}?
    \end{compactitem}
  \end{framed}
  \caption{Questionnaire on the effect of the inter-operator communication modalities we considered in our user study.}
  \label{fig-communication:questionnaire}
\end{figure}

\begin{itemize}
\item \textbf{Situational Awareness:} We measured situational awareness using the Situational Awareness Rating Technique (SART)~\cite{taylor2017situational} on a 4-point Likert scale~\cite{likert};
\item \textbf{Task Workload:} We used the NASA TLX~\cite{hart1988development} scale on a 4-point Likert scale to compare the perceived workload in each task;
\item \textbf{Trust:} We employed the trust questionnaire~\cite{uggirala2004measurement} on a 4-point Likert scale to assess trust in the system;
\item \textbf{Interaction:} We used a custom questionnaire on a 5-point Likert scale to quantify the effects of communication. The interaction questionnaire had the questions reported in Fig.~\ref{fig-communication:questionnaire}
\item \textbf{Performance:} We measured the performance achieved with each communication modality by using the points earned in each game.
\item \textbf{Usability:} We asked participants to select the features (text log, robot panel, object panel, and on-robot status) they used during the study. Additionally, we asked them to rank the communication modalities from 1 to 4, 1 being the highest rank. 
\end{itemize}

\begin{table}[h!]
\centering
\caption{Results with relationships between communication modes. The relationship are based on mean ranks obtained through Friedman's Test. The symbol $^*$ denotes significant difference ($p<0.05$) and the symbol $^{**}$ denotes marginally significant difference ($p<0.10$). The symbol $^-$ denotes negative scales and lower ranking is a good ranking.}
\renewcommand{\arraystretch}{1.3}
\begin{tabular}{c|c|c|c}
\hline
\textbf{Attributes}            & \textbf{Relationship}      & \textbf{$\chi^2(3)$}   & \textbf{$p$-value}    \\ \hline 
\multicolumn{4}{c}{\textbf{SART SUBJECTIVE SCALE}}                                                           \\ \hline\hline
Instability of Situation$^-$   & NC$>$IC$>$DC$>$MC$^{*}$    & $9.000$                 & $0.029$              \\ 
Complexity of Situation$^-$    & not significant            & $2.324$                 & $0.508$              \\
Variability of Situation$^-$   & IC$>$NC$>$MC$>$DC$^{*}$    & $9.303$                 & $0.026$              \\ 
Arousal                        & IC$>$NC$>$MC$>$DC$^{*}$    & $6.371$                 & $0.095$              \\
Concentration of Attention     & IC$>$NC$>$DC$>$MC$^{*}$    & $17.149$                & $0.001$              \\
Spare Mental Capacity          & not significant            & $5.858$                 & $0.119$              \\
Information Quantity           & MC$>$DC$>$IC$>$NC$^{*}$    & $15.075$                & $0.002$              \\
Information Quality            & MC$>$DC$>$IC$>$NC$^{*}$    & $15.005$                & $0.002$              \\
Familiarity with Situation     & not significant            & $6.468$                 & $0.101$              \\ \hline\hline
\multicolumn{4}{c}{\textbf{NASA TLX SUBJECTIVE SCALE}}                                                       \\ \hline\hline
Mental Demand$^-$              & not significant            & $2.226$                 & $0.527$              \\ 
Physical Demand$^-$            & not significant            & $2.165$                 & $0.539$              \\ 
Temporal Demand$^-$            & not significant            & $3.432$                 & $0.330$              \\ 
Performance$^-$                & not significant            & $0.412$                 & $0.938$              \\ 
Effort$^-$                     & not significant            & $1.450$                 & $0.694$              \\ 
Frustration$^-$                & not significant            & $4.454$                 & $0.216$              \\ \hline\hline
\multicolumn{4}{c}{\textbf{TRUST SUBJECTIVE SCALE}}                                                          \\ \hline\hline
Competence                     & not significant            & $4.740$                 & $0.192$              \\ 
Predictability                 & MC$>$IC$>$DC$>$NC$^{*}$    & $10.626$                & $0.014$              \\ 
Reliability                    & MC$>$IC$>$DC$>$NC$^{*}$    & $8.443$                 & $0.038$              \\ 
Faith                          & MC$>$IC$>$DC$>$NC$^{*}$    & $9.451$                 & $0.024$              \\ 
Overall Trust                  & MC$>$IC$>$DC$>$NC$^{**}$   & $6.633$                 & $0.085$              \\ 
Accuracy                       & not significant            & $1.891$                 & $0.595$              \\ \hline\hline
\multicolumn{4}{c}{\textbf{INTERACTION SUBJECTIVE SCALE}}                                                    \\ \hline\hline
Teammate's Intent              & DC$>$MC$>$IC$>$NC$^{*}$    & $19.610$                & $0.000$              \\ 
Teammate's Action              & MC$>$DC$>$IC$>$NC$^{*}$    & $13.810$                & $0.003$              \\ 
Task Progress                  & MC$>$DC$>$IC$>$NC$^{*}$    & $ 9.686$                & $0.021$              \\ 
Robot Status                   & not significant            & $ 0.811$                & $0.847$              \\ 
Information Clarity            & not significant            & $ 5.625$                & $0.131$              \\ \hline\hline
\multicolumn{4}{c}{\textbf{PERFORMANCE OBJECTIVE SCALE}}                                                     \\ \hline\hline
Points Scored                  & not significant            & $0.808$                 & $0.848$              \\ \hline 
\end{tabular}
\label{tab-communication:results}
\renewcommand{\arraystretch}{1}
\end{table}

\subsection{Results}
We analyzed the data using the Friedman test~\cite{friedman1937use} and summarized the results based on the significance and the mean ranks. Table~\ref{tab-communication:results} shows the summarized results along with relationship ranking between the communication modes. We formed the rankings for each scale using the mean ranks of the Friedman test. We categorized the relationship as significant for $p < 0.05$ and marginally significant for $p < 0.10$. 

\begin{figure}[t]
    \centering
    \includegraphics[width=0.49\textwidth]{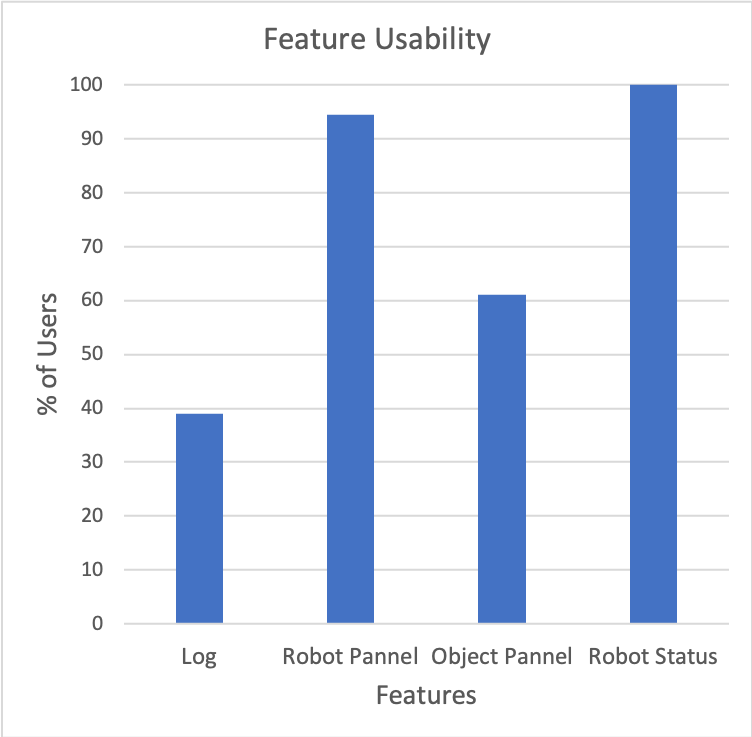}
    \caption{Feature Usability.}
    \label{fig-communication:feature_usability}
\end{figure}

\begin{figure}[t]
    \centering
    \includegraphics[width=0.49\textwidth]{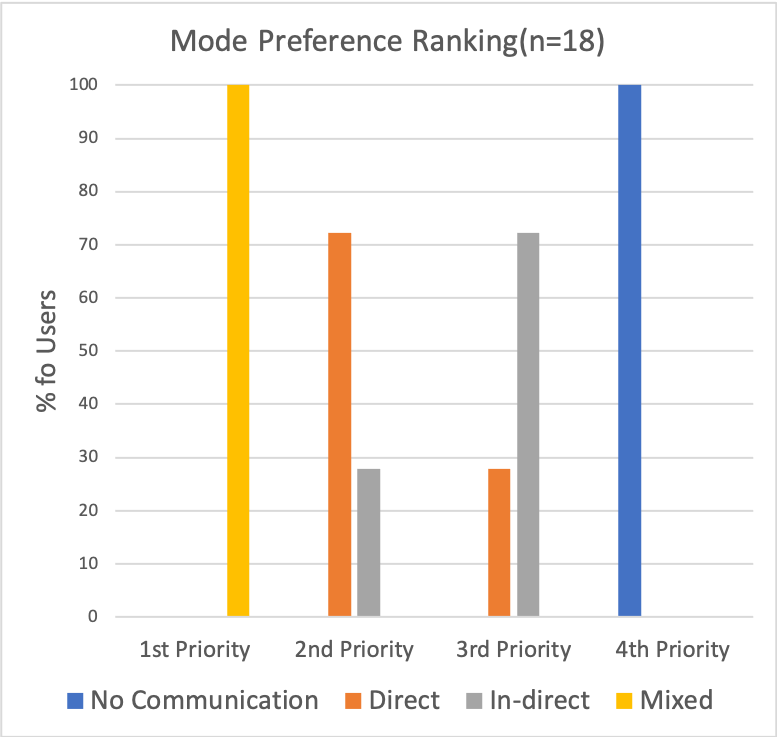}
    \caption{Task Preference.}
    \label{fig-communication:task_usability}
\end{figure}

Fig.~\ref{fig-communication:feature_usability} shows the usability results, i.e., the percentage of operators that used a particular feature to complete a specific task. Fig.~\ref{fig-communication:task_usability} shows how the participants ranked the communication modalities.

We used the Borda count~\cite{black1976partial} to quantify all the derived rankings based on data and preference to find an overall winner. We also inverted the ranking of the negative scales for consistency. Table~\ref{tab-communication:borda} shows the results.

\begin{table}[t]
\centering
\caption{Ranking scores based on the Borda count. The gray cells indicate the leading scenario for each type of ranking.}
\renewcommand{\arraystretch}{1.3}
\begin{tabular}{p{4cm}|c|c|c|c}
\hline\hline
Borda Count                                                         & NC   & DC    & IC    & MC                            \\ \hline\hline
Based on Collected Data Ranking (Table~\ref{tab-communication:results})           & 18   & 34    & 33    & \cellcolor[HTML]{EFEFEF}45    \\  
Based on Preference Data Ranking (Fig.~\ref{fig-communication:task_usability})    & 18   & 49    & 41    & \cellcolor[HTML]{EFEFEF}72    \\ \hline 
\end{tabular}
\label{tab-communication:borda}
\renewcommand{\arraystretch}{1}
\end{table}

\begin{figure}[t]
    \centering
    \includegraphics[width=0.49\textwidth]{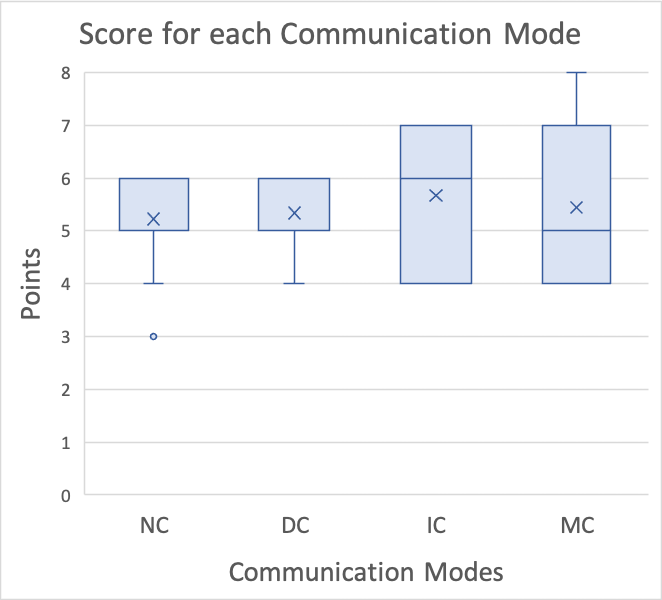}
    \caption{Task performance for each communication mode.}
    \label{fig-communication:performance}
\end{figure}

\begin{figure}[t]
    \centering
    \includegraphics[width=0.49\textwidth]{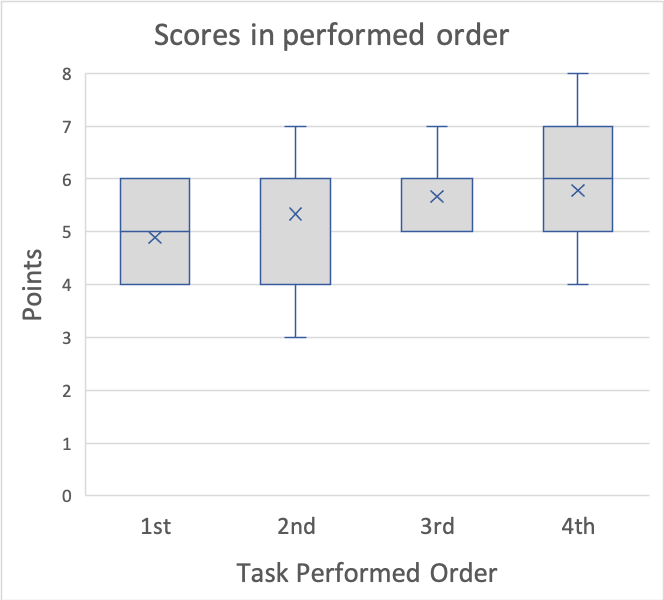}
    \caption{Learning effect in the user study.}
    \label{fig-communication:learning}
\end{figure}

\section{Discussion}
\label{sec-communication:discussion}

Table~\ref{tab-communication:results} shows that mixed communication (MC) has the best information quality and quantity, leading to the best awareness and trust. However, with more information, the operators experienced greater instability of situation and variability of situation (see the \textit{SART Subjective Scale} section of Table~\ref{tab-communication:results}). The participants indicated mixed communication as the best choice in both the data and their  expressed preference (see Table~\ref{tab-communication:borda}), confirming the hypotheses \textbf{H1} and \textbf{H2}. 

Direct communication (DC), compared to indirect communication (IC), had better information quality and quantity, leading to better awareness. However, similar to mixed communication, the operators experienced greater instability of situation and variability of situation. Although trust is higher in indirect communication, operators prefer direct communication. The Borda count for preference data shows direct communication ranks better than indirect communication, supporting hypothesis \textbf{H3}. 

However, in the absence of direct communication, the operators concentrated more on the task, leading to a higher level of arousal and better trust (see \textit{Trust Subjective Scale} section of Table~\ref{tab-communication:results}). Although indirect communication provides more diverse and more visually augmented information, operators relied on direct communication when working as a team. We also observed operators directly communicating either at the start of the task, to define a strategy, or near the end of the task, to coordinate the last part of the task. One reason can be familiarity of information. Humans are more familiar with direct communication. The participants were new to the interface and were unable to use it as effectively as communicating directly. This raises another research question and a potential future study on comparing the effects of mixed communication on novice operators and on expert operators.

Our experiments did not detect a substantial difference in performance across communication modes. However, we hypothesize that this lack of difference is because of the learning effect across the trials each team had to perform. Fig.~\ref{fig-communication:performance} represent the points earned for each communication mode and Fig.~\ref{fig-communication:learning} shows the learning effects as the increase in objective performance in order of the performed task. 

\section{Conclusion and Future Work}
\label{sec-communication:conclusion}
We presented a study on the effects of inter-operator communication in multi-human multi-robot interaction. We concentrated on two broad types of communication: \emph{direct}, whereby two operators engage in an exchange of information verbally or non-verbally (e.g., with gestures, body language); and \emph{indirect}, whereby information is exchanged through a suitably designed interface. We presented the design of our interface, which offers multi-granularity control of teams of robots.

In a user study involving 18 participants and 9 robots, we assessed the effect of inter-operator communication by comparing direct, indirect, mixed, and no communication. We considered metrics such as awareness, trust, workload and usability. Our results suggest that allowing for mixed communication is the best approach, because it fosters flexible and intuitive collaboration among operators.

In future work, we will study how training affects communication. In comparing direct and indirect communication, we noticed that our novice participants preferred direct to indirect communication. Our experience as online videogamers leads us to hypothesize that novice users heavily lean on verbal communication due to a lack of experience with the information conveyed by the interface. As the experience of the operators increases, we expect direct communication to become more sporadic and high-level. At the same time, more expert users might request advanced features to further make indirect communication more efficient.

\section*{Acknowledgements}
This work was funded by an Amazon Research Award.

\bibliographystyle{IEEEtran}
\bibliography{ref}

\end{document}